\documentclass[preprint,12pt]{elsarticle}
\usepackage{graphicx}
\usepackage{epsf}

\usepackage{amssymb}

\journal{Journal of Systems and Software}

\begin{document}

\begin{frontmatter}



\title{Addressing the non-functional requirements of computer vision systems: A case study}

\author[label1]{Shannon Fenn}
\author[label1]{Alexandre Mendes}
\author[label2]{David Budden}

\address[label1]{School of Electrical Engineering and Computer Science,\\
              Faculty of Engineering and Built Environment,\\
              The University of Newcastle, Callaghan, NSW, 2308, Australia.\\
              \texttt{\{Shannon.Fenn,Alexandre.Mendes\}@newcastle.edu.au}\\}
							
\address[label2]{Systems Biology Laboratory,\\
              Melbourne School of Engineering,\\
              The University of Melbourne, Parkville, VIC, 3010, Australia.\\
              \texttt{David.Budden@unimelb.com.au}\\}

\begin{abstract}
Computer vision plays a major role in the robotics industry, where vision data is frequently used for navigation and high-level decision making. Although there is significant research in algorithms and functional requirements, there is a comparative lack of emphasis on how best to map these abstract concepts onto an appropriate software architecture.

In this study, we distinguish between the functional and non-functional requirements of a computer vision system. Using a RoboCup humanoid robot system as a case study, we propose and develop a software architecture that fulfills the latter criteria.

The modifiability of the proposed architecture is demonstrated by detailing a number of feature detection algorithms and emphasizing which aspects of the underlying framework were modified to support their integration. To demonstrate portability, we port our vision system (designed for an application-specific DARwIn-OP humanoid robot) to a general-purpose, Raspberry Pi computer. We evaluate performance on both platforms and compare them to a vision system optimised for functional requirements only.

The architecture and implementation presented in this study provide a highly generalisable framework for computer vision system design that is of particular benefit in research and development, competition and other environments in which rapid system evolution is necessary.

\end{abstract}


\end{frontmatter}


\section{Introduction}
\label{intro}

There is currently significant research into developing and optimising the functional requirements (\emph{e.g.}, feature detection and classification) of computer vision systems, both within the multi-billion dollar robotics industry~\cite{Chen_2011,Kapach_2012,Budden_2013,Sonka_2014} and the more general field of industrial automation~\cite{Wu_2013,Wang_2011,Aldrich_2010}. Many of these application domains lend themselves to the initial development of a highly-specialised computer vision system, without the need for modifiability, extensibility or portability as design considerations. However, other domains (\emph{e.g.}, robot soccer) have particular need to foster innovation in these additional non-functional requirements, due to the rapid evolution of both cutting-edge hardware and the physical domain itself~\cite{Kitano_1997}. Despite the clear importance of these requirements, there is currently a notable lack of research and practical guidelines on how best to map them onto an appropriate software architecture.

In this study, we propose a two-layered architecture with \textit{interface}, \textit{control} and \textit{data storage} components in the top layer; plus all the components for \textit{feature detection} in the bottom layer. The system promotes \textit{hardware independence} by having a specific interface component to communicate with the camera and other sensors; \textit{modifiability} by adopting a modular approach for the implementation of the algorithms; and \textit{efficiency}, with the introduction of a controller module to use those algorithms in a more intelligent fashion. Using the vision system of the NUbots humanoid RoboCup team~\cite{Annable_2013,Kulk_2012}\footnote{\texttt{\scriptsize{https://github.com/shannonfenn/Multi-Platform-Vision-System}}} as a case study, we demonstrate how the proposed architecture both:

\begin{itemize}
\item addresses the non-functional requirements of modifiability, extensibility and portability; and
\item enables the rapid development and integration of feature detection algorithms to better address the functional requirements of a rapidly-evolving domain (\emph{e.g.}, robot soccer).
\end{itemize}

This paper is organized as follows. In Section~\ref{sec:platforms} we describe the two platforms used for evaluating the proposed architecture. Section~\ref{sec:visionsystemdescription} describes the designs of both original and new vision systems, both
in terms of functional (algorithms) and non-functional (architecture) requirements. In Section~\ref{sec:experimentalDesign} we describe the image data and the algorithms for feature detection used in our tests. Then, in Section~\ref{sec:results} we present the computational results for performance related to the functional requirements for the system. In Section~\ref{sec:porting} we explain the porting of the new system from the DARwIn-OP to a Raspberry Pi platform; followed by a discussion of our main findings and their implications in computer vision system design.

\section{The DARwIn-OP and the Raspberry Pi platforms}
\label{sec:platforms}

Portability is one of the key non-functional requirements for software systems designed to be used in hardware dependent environments. Continuous progress in robotics frequently raises the difficult question of when to change platforms to better take advantage of such advances~\cite{Smith_2004}. For instance, the University of Newcastle robotic soccer team has changed platforms twice since 2002 -- first, from the Sony Aibo\footnote{\texttt{http://www.sonydigital-link.com/AIBO/}} to the Aldebaran NAO\footnote{\texttt{http://www.aldebaran-robotics.com/en/}}, and then to the Robotis DARwIn-OP\footnote{\texttt{http://www.robotis.com/xe/darwin\_en}}. Each of those required a major overhaul of the entire software system; mainly due to changes in sensors, motors, and the adaptation of higher level procedures such as walking and kicking, which are platform-specific.

The vision system described in this paper was designed for utilisation on multiple platforms with as little re-design as possible. In order to demonstrate portability, it was tested with two platforms: the DARwIn-OP and the Raspberry Pi. The DARwIn-OP features a 1.6 GHz Intel Atom Z530 with 4 GB of RAM, Ubuntu operating system, a Logitech C905 camera, plus actuators, sensors, etc. The main requirement on the vision system imposed by the platform was efficiency, due to the low-power, single-core processor. Two other main requirements were flexibility and modifiability. The system will be used in future RoboCup competitions, and game rules typically change every year to make the matches more realistic. Historically, most of those changes impact either vision or behavior.

The second platform tested was a Raspberry Pi, which is a credit card-sized, single-board computer. It has an ARM1176JZF-S 700 MHz processor with 512 MB of RAM and Ubuntu operating system. It can be connected to a standard USB webcam, keyboard and monitor, and was developed mainly for educational purposes\footnote{\texttt{http://www.raspberrypi.org/}}. Porting to a Raspberry Pi platform creates a few challenges from the software architecture perspective. Due to the lack of sensors and motors attached to it, the vision system has to be adapted to obtain that data from so-called \textit{mock components} that will mimic the behaviour of the real hardware. Further, since the system must be be moved without its partner components (Localisation, Behaviour and Locomotion) the usual external communication infrastructure is missing. The less modifications required, the easier the porting process.

\section{Original and new vision systems architectures}
\label{sec:visionsystemdescription}
Software architecture is the high level description of a software system. Its main goal is to describe the system's components and their interactions; how functional requirements map onto them; and how non-functional requirements are promoted. The design of the software architecture is arguably the first step in the software design process, preceding the definition of classes and data structures and, of course, the coding process itself~\cite{Gordon_2006,Taylor_2009}. In this section we present the architecture diagrams and overviews of the original and new vision systems. We compare their main features and justify the pros and cons of the design decisions in each of them.

\subsection{Original vision system}
The original vision system was first designed in 2002 and implements a pipeline architecture, due to the highly inter-dependent nature of the various components. Since then, it has undergone several improvements in terms of functional requirements, but the architecture remained the same. A diagram of its main components is shown in Figure~\ref{fig:OriginalSystem}. The inputs to the system come from three sources:

\begin{figure}[b!]
\center
  \includegraphics[trim=4cm 4.5cm 3.5cm 4.0cm, clip=true, width=0.55\textwidth]{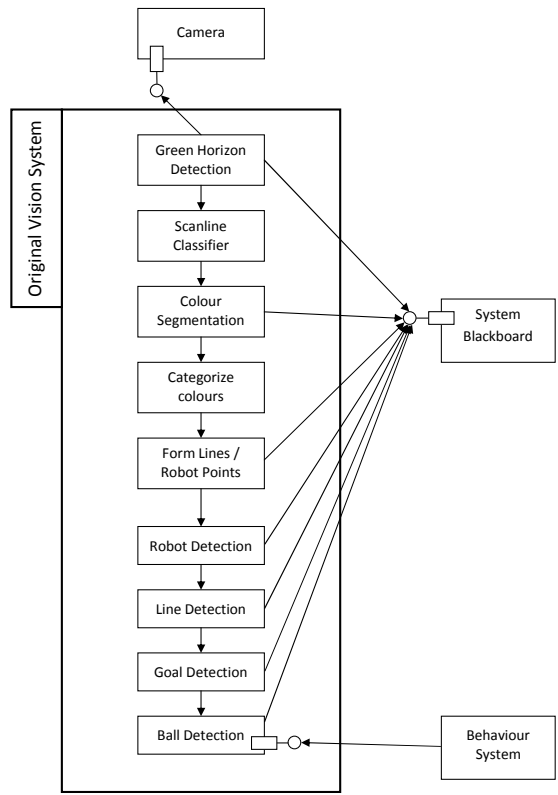}
\caption{Original vision system architecture design: The architectural style is a pipeline, due to the highly inter-dependency
nature of its components. Even though it promotes efficiency by minimising the overhead that would be present
in a more robust architecture, it becomes inflexible as it forces the sequential application of every object
detector. Notice also the high-dependency on the external System Blackboard module - 7 out of 9 modules rely on its interface.}
\label{fig:OriginalSystem}       
\end{figure}

\begin{itemize}
\item The \emph{image stream} from the camera, represented by the external component `Camera';
\item \emph{Kinematics data} from the Locomotion system, stored in the component `System Blackboard';
\item The \emph{colour classification lookup table}, also stored in the System Blackboard.
\end{itemize}

The sole requirement of the vision system is to take these inputs and determine the relative locations of predefined field objects; \emph{e.g.}, ball and goalposts. That information will then be accessed by the Behaviour system to determine the next action the robot should take. The pipeline design was useful as it allowed for the maximum efficiency by minimising the overhead that would be present in a more robust architecture. It is however fairly inflexible and does not readily allow for parallel processing, which will become a reality when platform power demands decrease and multi-core chipsets are used. Another significant drawback of the original system is that it enforces the application of every processing stage, for every frame. This translates into unnecessary processing in situations where, for instance, there are only one or two features of interest (\emph{e.g.}, ball and goals). In addition, the lack of a data wrapper created the dependency of 7 out of 9 modules, on the System Blackboard interface, as can be seen in Figure~\ref{fig:OriginalSystem}.

\subsection{New vision system}
\begin{figure}[h!]
\center
  \includegraphics[trim=4cm 5.5cm 3cm 5cm, clip=true, width=0.6\textwidth]{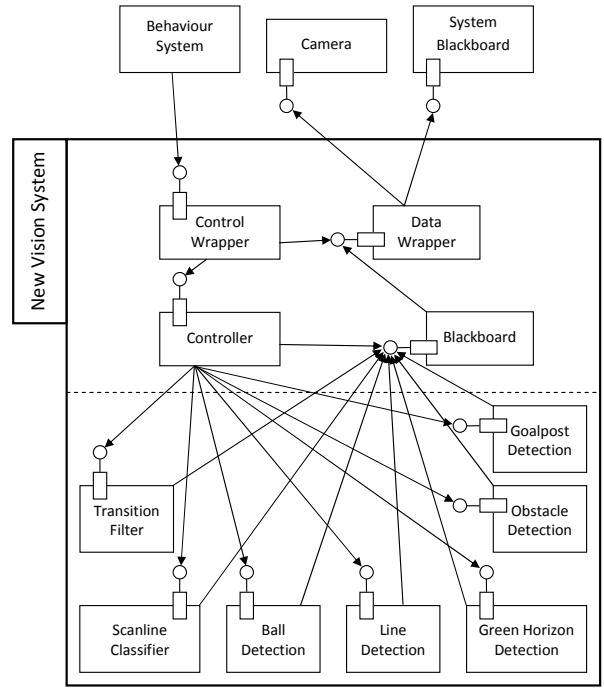}
\caption{New vision system architecture design: Notice the two-layer system, with the top layer concentrating all
external communication into two wrapper modules. The top layer also contains a Controller module that accesses the bottom
layer's processing stages; and a Blackboard module that stores locally all information required by those processing stages, as well as
their results. The system is considerably more flexible, easier to maintain and to modify. It also allows for the Controller
module to determine which processing stages should be applied, resulting in better flexibility.}
\label{fig:NewSystem}       
\end{figure}

The new vision system was designed with a focus on modularity, portability and modifiability (see Figure~\ref{fig:NewSystem}). Vision has to communicate with three external components - Behaviour System, Camera and System Blackboard. As the name indicates, the Behaviour system determines how the robot should behave and which action to take next. Camera represents the robot's on-board camera. The System Blackboard module contains information about the environment and the robot itself; and can be accessed by any of the other three systems of the robot's software, namely \textit{Behaviour}, \textit{Locomotion} and \textit{Localisation}.

Communication between the vision system and the three external components in the new design is achieved by the Control Wrapper and Data Wrapper modules. The Control Wrapper is an external interface and allows Behaviour to access and control several elements of the vision system. By designing this interface through a wrapper we trade a very slight decrease in efficiency (through a small number of redundant method calls) for a significant increase in modifiability. This modifiability comes as the Control Wrapper is the only module that needs to change for the system to be invoked in any different way (that is, provided the change does not require the vision system's internal \emph{behaviour} to be extensively modified).

The Data Wrapper allows the vision system to access the external data it needs (image frames, kinematic data, etc). The same reasoning for modelling the control interface as a wrapper is applicable for the data interface - small reductions in efficiency allow for a more modifiable design. This approach also allows for data validity to be maintained as simply as possible. The Data Wrapper is responsible for maintaining a copy of the relevant data accessed at the same time as the current frame, insulating the system against poorly maintained external data structures. For instance, if any sensor data is updated on the System Blackboard by, say, the locomotion system whilst the vision system is still analysing a frame, the Data Wrapper still has a local copy. This decouples the data responsibilities of the various systems and ensures Vision is as robust as possible to external changes.

Having these interfaces as wrappers has one extra advantage: rapid testing. Either wrapper can be replaced by a stub providing dummy data, invoking the vision system with extra debugging information. This also allowed for the system to be quickly integrated into the Raspberry Pi platform, interfacing with a stand-alone web camera, without kinematic or sensor data, in a near seamless fashion. Indeed, most of the vision system implementation and testing was made on a desktop PC, without using the real robotic platform. However, it is important to emphasize that the computational tests numbers presented in this paper indeed reflect the performance of the system on the DARwIn-OP and Raspberry Pi platforms.

The Controller module calls the processing stages in order; and stages can be optionally ignored; \emph{e.g.}, ball or goalpost detection. This allows for current state knowledge to increase the efficiency of the system. For example, consider the situation where the Localisation system has a high certainty of the robot location, but a low certainty of the ball location. In that case, it does not need any landmark information to determine where the robot is; and the vision's priority becomes simply to identify and locate the ball as quick as possible. By suppressing goalpost and line detections, the vision system reduces its processing time, freeing up resources for other parts of the software.

The Blackboard module was implemented following the singleton pattern~\cite{Yacoub_2003}. This was for two reasons: the blackboard is a single object throughout the vision system and it should not be possible for multiple copies to be instantiated; and it allows for simple access in the form of a static access method, improving performance. The Blackboard module is a central repository of current state information, specifically for the vision system; and each processing stage can request any input information needed from it and posts results back. This increases the complexity of the system somewhat by requiring correct process order to be maintained by a controller, but allows for rapid reordering of independent stages and incorporation of new stages.

In addition to the two wrappers, and Blackboard and Controller modules, there are seven processing modules, namely Scanline Classifier, Transition Filter, Line Detection, Green Horizon Detection, Ball Detection, Goalpost Detection and Obstacle Detection. Those are called sequentially by the Controller to carry out the tasks necessary to identify elements and features in the soccer field and allow the robot to play correctly.

\section{Experimental design}
\label{sec:experimentalDesign}
In this section, we describe the dataset used to compare the two vision systems and the algorithms for feature detection. The image stream was a collage of five typical playing scenarios - two of them were recorded in a laboratory environment (with a full indoor field), and three were recorded at previous RoboCup competitions. The total time is 5 minutes - equivalent to 9,000 image frames. Each of the five image streams is continuous and was taken at 30 frames-per-second, exactly the same condition of a real soccer match. All elements normally observed in a match, such as the ball, field lines, other robots and goalposts are present. The five scenarios along with a short description is given next:

\begin{itemize}
\item \textit{Lab 1}: Controlled environment at the lab. Occasional variable lighting and image quality. Background noise is present but minimal field occlusion.
\item \textit{Lab 2}: Controlled environment at the lab, taken after lighting change to a more stable condition. There is little to no variability in image quality or lighting, and no occlusion.
\item \textit{Difficult}: Stream with poor lighting, poor image quality and significant field occlusion, where objects of interest are difficult, or impossible to classify correctly. The stream was taken on one of the fields at RoboCup 2012, in Mexico City.
\item \textit{RC 2012}: This stream was also recorded during RoboCup 2012, on a field with a prominent lighting intensity gradient from one end to the other. The lighting variability complicates the task of feature classification. Varying image quality and background noise.
\item \textit{RC 2013}: Stream recorded during RoboCup 2013, in The Netherlands. Predominantly high quality images with consistent lighting. Also contains a large number of opponent robots and human obstacles; \emph{e.g.}, referees and team captains.
\end{itemize}


\subsection{Green horizon detection}
\label{sec:greenhorizon}

\begin{figure}[h!]
\center
  \includegraphics[width=0.59\textwidth]{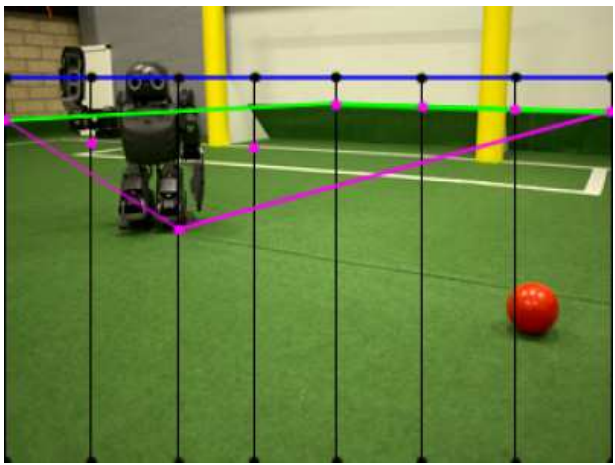}
\caption{Camera frame and green horizon detection results: The image is a real frame captured by the DARwIn-OP onboard camera of the playing field. It shows several important features - ball, goalposts, another robot and several field lines. Notice that all those features can still be detected if we consider only the part of the image below the so-called green horizon; \emph{i.e.}, the point where the field ends. The blue line indicates the true horizon, obtained from kinematics information about the robot's head angle. The vertical lines are scanlines used to determine where the green pixels start (marked by the magenta markers). Finally, the green horizon is calculated as the top segments of the convex hull associated to those markers.}
\label{fig:GreenHorizon}       
\end{figure}

Green horizon detection is a procedure used to improve the system's performance by determining which parts of the image are relevant or not. All objects of interest are within the field, which has a green colour; everything else can be considered noise for soccer playing purposes \emph{e.g.}, walls, crowds, etc. (see Figure~\ref{fig:GreenHorizon}). 

The only objects of interest which might extend above the green horizon are the goalposts, and they are detected using a less computationally intensive method (see Sections~\ref{sec:scanlines} and~\ref{sec:goalposts}). The procedure to determine the green horizon is as follows. First, it requests the location of the real horizon in screen coordinates, calculated using a forward kinematics chain model and a camera projection, from the external Blackboard module. The method then scans the image downwards from this horizon using evenly-spaced, vertical scan-lines to determine where the first green pixels start - and stores those points (indicated with magenta markers in Figure~\ref{fig:GreenHorizon}). The green horizon is calculated as the upper convex hull formed from those points. The procedure is very fast ($ \mathcal{O}\left(n\right) $) and robust to features blocking parts of the horizon (\emph{e.g.}, the robot in Figure~\ref{fig:GreenHorizon}), as well as being capable of handling the changing geometry of the green border in the camera frame.

\subsection{Scanline classifier and colour segmentation}
\label{sec:scanlines}

\begin{figure}[b!]
\center
  \includegraphics[trim=1cm 1cm 1cm 1.5cm, clip=true, width=0.46\textwidth, angle=-90]{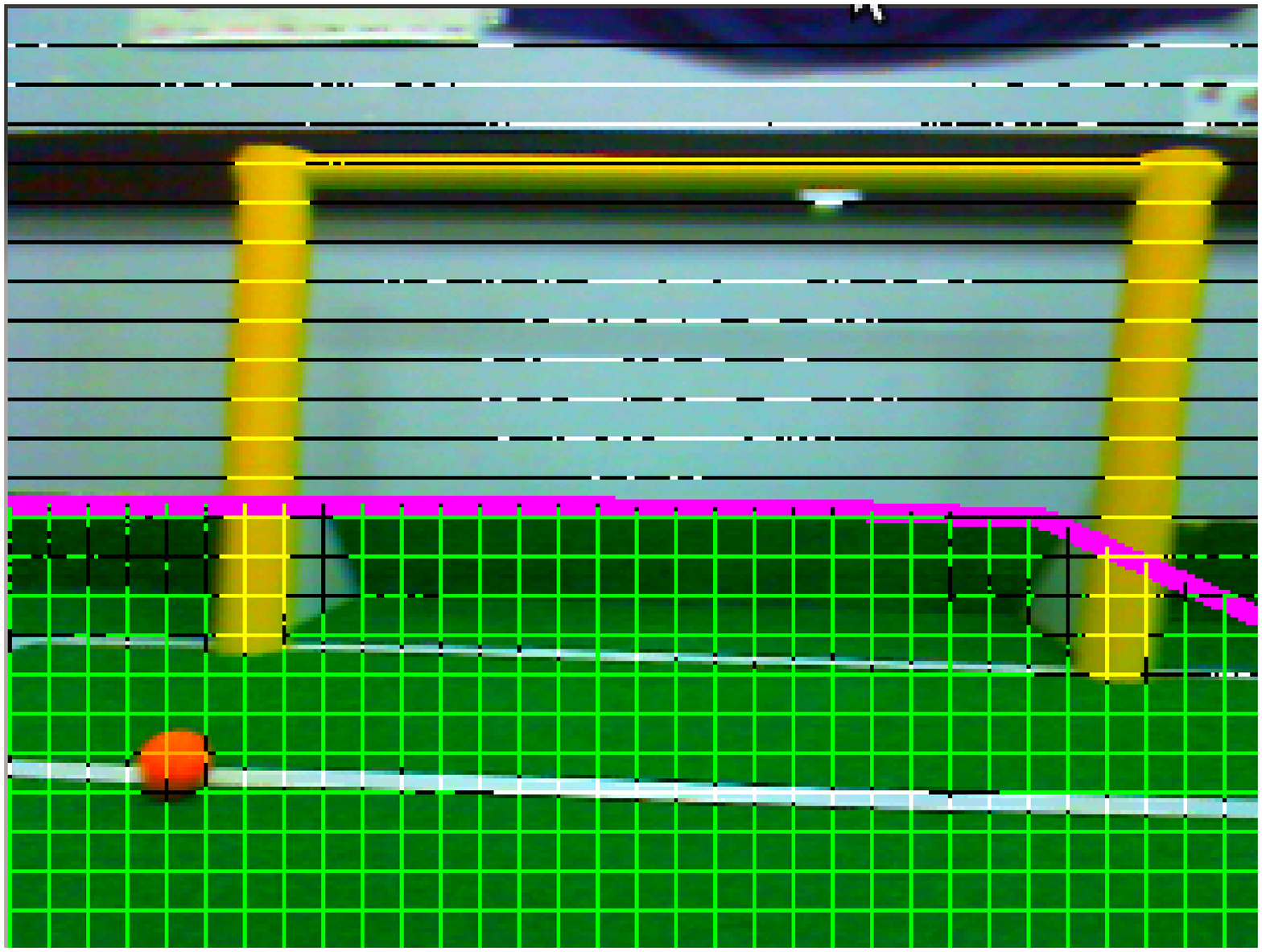}
\caption{Scanline classifier and colour segmentation are used as a pre-processing step for the detection algorithms. Vertical and horizontal scan lines are considered for sequential pixels of the same colour, which are indicative of objects of interest. In the image, the green horizon is represented by the magenta line, and scanlines are used to detect the ball, goalposts and field lines. In order to reduce computational time, the area below the green horizon uses both vertical and horizontal scanlines for better accuracy. The area above the green horizon is scanned using horizontal lines only, and only if there was a goalpost detected below the horizon.}
\label{fig:Scanlines}       
\end{figure}

This stage is a pre-process to all object detection procedures. It comprises an image scan followed by a segmentation for specific colour sections~\cite{Budden_2013sal}. Due to time constraints we do not segment the entire image. Instead, we separate the image into below/above green horizon areas. For the area below the green horizon, vertical and horizontal scan lines, evenly-spaced by a small number of pixels, are considered for sequential pixels of the same colour. These form segments which are the basis for all further filtering and detection, and are stored in a special data structure in the vision system's Blackboard module. For the area of the image above green horizon, we only use horizontal scanlines, as the only features of interest in that region are the goalposts. Figure~\ref{fig:Scanlines} shows the result of the scanline and colour segmentation procedures applied to a typical image of the playing field.

\subsection{Feature detection}
Colour transitions form the input to all feature detection algorithms except for obstacle (\emph{e.g.}, robot) detection. Each of these processes take the transitions which have been mapped to their corresponding field objects, and apply a combination of clustering, filtering and model-fitting techniques to identify valid objects.

\subsubsection{Ball}
Ball detection uses the geometric mean of the pixels' locations for all transitions that might represent a ball; followed by a 4-point occlusion detection method. Valid transitions representing a ball are ``\emph{orange$\rightarrow$all colours}" or ``\emph{all colours$\rightarrow$orange}". The method is very efficient; accurate in terms of distance-to-ball estimation; and robust to partial occlusion (up to 50\%) of the ball~\cite{Budden_2012}.

\subsubsection{Goalposts}
\label{sec:goalposts}
Goals are integral to the localisation of the robot. They are large, static and monochromatic features; and therefore are relatively simple to extract. Valid transitions representing a goalpost are ``\emph{yellow$\rightarrow$all colours}" or ``\emph{all colours$\rightarrow$yellow}".

The statistical model fitting method RANdom Sample And Consensus (RANSAC) is used to fit lines (further detail in Section~\ref{sec:fieldlines}) to the filtered transitions - thus detecting the vertical edges of the post which are typically very clear. That step is then followed by a simple heuristic to pair the lines into post objects~\cite{Flannery_2014}. This procedure is able to reliably handle images with no posts and goal-coloured noise in the background as well as those with one or two posts in the frame.

\subsubsection{Obstacles}
Obstacles are found by considering breaks in the green horizon. The first green pixel markers used for constructing the green horizon that are not in the convex hull are sent to list of obstacle candidates. For instance, in Figure~\ref{fig:GreenHorizon}, those would be markers 2, 3 and 4, counted from left to right. If there are at least $\alpha$ consecutive scans where the markers do not belong to the green horizon, then the set of scans forms a potential obstacle. The width is determined from the outer scanlines, and the base of the obstacle from the lowest scan.

\subsubsection{Field lines}
\label{sec:fieldlines}
Field lines are an important source of information for the robot to localise itself in the field. They are visible from most field positions, as opposed to landmarks that can often be absent due to the small camera field of view. Any ``\emph{white$\rightarrow$all colours}" or ``\emph{all colours$\rightarrow$white}" transition might represent a line border. The vision system uses the RANSAC algorithm mentioned in Section~\ref{sec:goalposts}~\cite{Flannery_2014} due to its resilience to noise and, in particular, extreme numbers of outliers; which makes it effective for finding multiple models within a single dataset. For more information about additional computer vision methods and an overview of related algorithms applied to robotic soccer, we refer the reader to reference~\cite{Nadarajah_2013}.

\section{Computational results}
\label{sec:results}

This section is divided into two parts. The first one describes a performance comparison between original and new versions of the vision system. The second part explains the process to port the new vision system from the DARwIn-OP to the Raspberry Pi platform.

\subsection{Performance comparison}
To conduct a performance comparison between the two versions of the vision system, we ran a series of tests using the dataset described in Section~\ref{sec:experimentalDesign} on the DARwIn-OP platform. The goal was to measure the average time to process each image frame, thus determining if the increase in flexibility and portability came at the cost of reduced efficiency. The mean and standard deviation of the processing times for each image frame are shown in Table~\ref{tab:processingTimes}.

\begin{table}[h!]
\center
\caption{Performance comparison between the original and the new vision systems. The figures represent the average total time (standard deviations are in subscript, between brackets) required to process a single frame on the DARwIn-OP platform. Notice that the processing time was already well below 30ms, indicating that the original system did achieve above real-time image processing. However, the new system is 54\% faster, primarily due to better data handling and reduced overhead.}
\label{tab:processingTimes}       
\small
\begin{tabular}{llll}
\hline\noalign{\smallskip}
              &    & Original vision system  & New vision system\\
Image stream  &  Number of frames     & in ms per frame       & in ms per frame\\
\noalign{\smallskip}\hline\noalign{\smallskip}
Lab 1      & $5090$      & $18.73_{(3.89)}$ &	$13.37_{(3.04)}$   \\
Lab 2	     & $ 470$      & $25.95_{(4.63)}$ &	$ 8.29_{(2.35)}$   \\
Difficult	 & $ 175$      & $ 9.68_{(1.70)}$ &	$ 7.21_{(1.15)}$   \\
RC 2012	   & $2640$      & $20.31_{(5.32)}$ &	$12.56_{(2.81)}$   \\
RC 2013		 & $ 625$      & $18.97_{(3.58)}$ &	$10.04_{(2.09)}$   \\
\noalign{\smallskip}\hline\noalign{\smallskip}
           & $9000$      & $19.41_{(4.85)}$ & $12.52_{(3.24)}$ \\
\noalign{\smallskip}\hline
\end{tabular}
\end{table}

The results show an average reduction of 35.5\% in processing time between original and new vision systems. The original vision system was already capable of faster than real-time image processing, achieving a mean frame rate of 52 fps. The new system increased that frame rate to 80 fps. The improvement in the performance resulted from three sources, listed below.

\begin{itemize}
\item A better architecture design that reduced communication overhead;
\item A focus on intelligent data reduction in earlier processing stages to reduce the input size to later, more complex algorithms, and;
\item The principled application of specialised data structures to store information about the several objects needed during the several stages of image processing and object detection.
\end{itemize}

\subsection{Porting to the Raspberry Pi platform}
\label{sec:porting}
A port from the DARwIn-OP to the Raspberry Pi platform was conducted to demonstrate the portability and flexibility of the new vision system. To emulate the behaviour of the DARwIn-OP platform in the Raspberry Pi, a modification \textit{external to the new vision system} had to be made.

For results to be comparable, modifications to the camera access section of the Data Wrapper were made to provide the vision system with pre-recorded frames captured on the DARwIn-OP platform. This same modification was made to the version run on the DARwIn-OP and a similar change was made to the original system - so that the comparison between implementations was fair.

In addition to that external change, the internal changes were as follows:

\begin{itemize}
\item The Data Wrapper class was modified to provide the system with `dummy' kinematics data and other system data, as there was no physical robot with motors, camera or other sensors attached to the Raspberry Pi. This required less than a fifth of the methods to be modified or removed.

\item The Control Wrapper class was modified so that the system was more easily used as a stand-alone process instead of as a submodule. As this is a relatively small interface, all seven methods were modified or removed in the transition.
\end{itemize}

The only difference between the DARwIn-OP and the Raspberry Pi test results was the performance, as expected, due to the different processors in the platforms. Beyond the reduced CPU capacity, the Raspberry Pi also has a number of other limiting factors including possessing only a quarter of the system memory of the DARwIn-OP. The new vision system running in the Raspberry Pi achieved an average frame rate of 23 fps (see Table~\ref{tab:processingTimesRPi}), over a period of 5 minutes of image capturing and processing (9000 frames). Compared to the 80 fps rate obtained in the DARwIn-OP platform, that represents a decrease of 70\%, in line with the processing power difference between the processors. The accuracy and reliability of the new vision system remained the same, as expected, since there were no changes in functionality. Porting the whole new vision system took \textit{less than 6 hours of re-implementation and testing}, and affected only two of its modules: the Data Wrapper and Control Wrapper modules. The Data Wrapper had 101 lines of code changed (53\% of the code), in 12 of 34 methods; and the Control Wrapper had 17 lines of code changed (59\% of the code), in 5 of 5 methods.

\begin{table}[h!]
\center
\caption{Performance comparison between the new vision system implementation on the DARwIn-OP and Raspberry Pi platforms. Similarly to Table~\ref{tab:processingTimes}, the figures represent the average total time (standard deviations are in subscript, between brackets) required to process a single frame.}
\label{tab:processingTimesRPi}       
\small
\begin{tabular}{llll}
\noalign{\smallskip}
\noalign{\smallskip}
\hline\noalign{\smallskip}
              &              & DARwIn-OP   & Raspberry Pi  \\
							&               & performance in & performance in\\
Image stream  &  Number of frames      & ms per frame      & ms per frame \\
\noalign{\smallskip}\hline\noalign{\smallskip}		
Lab	1			 & $5090$    & $13.37_{(3.04)}$  & $44.52_{( 7.65)}$ \\
Lab 2	     & $ 470$    & $ 8.29_{(2.35)}$  & $47.01_{(12.18)}$ \\
Difficult	 & $ 175$    & $ 7.21_{(1.15)}$  & $12.11_{( 1.41)}$ \\
RC 2012	   & $2640$    & $12.56_{(2.81)}$  & $47.94_{(16.89)}$ \\
RC 2013	   & $ 625$    & $10.04_{(2.09)}$  & $28.04_{( 4.16)}$ \\
\noalign{\smallskip}\hline\noalign{\smallskip}
           & $9000$    & $12.52_{(3.24)}$   & $43.88_{(12.97)}$ \\
\noalign{\smallskip}\hline
\end{tabular}
\end{table}

A large factor in these changes was the replacement of code which accessed the external system blackboard or communicated with the Localisation system, with code to generate mock data and communicate with the Raspberry Pi peripherals. For example, the method that provides the system with an artificial horizon based on kinematic data (see Section~\ref{sec:greenhorizon}), was replaced with a stub that provided a hard-coded horizon at the top of the image frame (so that the image was processed fully, regardless of the declination of the camera during capture). After this experiment, it becomes clear that when the vision system is ported to a new, more advanced platform in the future, it should be a relatively straightforward task.

\section{Discussion}
\label{sec:discussion}
Whilst the system, as presented, is tailored for soccer scenarios, it is a generic architecture that can support more complex settings; \emph{e.g.}, multiple feature detectors, either simple and self-contained modules, or complex subsystems comprised of several independent stages. Any colour-based design can also take advantage of the existing segment and transition filter pre-processing components.

It should be noted that there has been no leveraging of parallelism within this design. Parallelism, as an oft-noted inherent aspect of vision processing, could be well utilised within the architecture for significant performance gains - both fine grained within individual algorithms, as well as course grained between independent modules. This would form a valuable direction for enhancing the architecture, allowing the use of more complex detection algorithms as higher performance parallel computing becomes more available to low power platforms.

In addition to these non-functional improvements, Table~\ref{tab:processingTimes} shows that the new system markedly outperforms the previous system in multiple scenarios, including competition venues, intentionally noisy images and the laboratory environment. As computational resources on mobile platforms are restricted, this efficiency increase can pose a significant gain in the functional performance of the robot.

This design is not without its drawbacks though. The flexibility of the internal blackboard pattern allows for coding mistakes during modification of the controller to break the system flow (as there is no dependency constraint in the design). Future modifications to the system need to be careful to ensure the data expected by a later-stage module is provided to the blackboard in the expected form.

A further drawback is the use of singleton classes. These were included in order to reduce the communication overhead involved in passing references to the internal Blackboard and Data Wrapper objects, by instead having them accessible via a static \emph{getInstance()} method. This choice however means that utilising multiple cameras with this system will require architectural redesign or careful data management as simply instantiating two distinct versions of the system is not possible. This is a potential direction for further improvement.

The new vision system and the five image streams are available for download as supplementary material\footnote{\texttt{https://github.com/shannonfenn/Multi-Platform-Vision-System}}.

\section{Conclusion}

In this study, we have proposed, implemented and evaluated a software architecture design for general computer vision systems to address the non-functional requirements of modifiability, extensibility and portability. In addition to demonstrating the fulfillment of these criteria, we further demonstrate that a computer vision system optimised for non-functional requirements can yield improved performance (quantified as execution time to process image frames).

We selected RoboCup humanoid robot soccer as an example of a domain subject to rapidly-evolving, domain-specific requirements, in which non-functional requirements are particularly critical. The vision system detailed in this study has been successfully applied by two-time world champion RoboCup team NUbots during the 2013 (Netherlands) and 2014 (Brazil) competitions\footnote{\texttt{http://www.robocup.org}}. We propose that the underlying software architecture is readily extensible to a far wider range of computer vision domains, and moreover, that the complementary development of systems to address both functional and non-functional requirements is critical to future computer vision research.

\end{document}